\documentclass[11pt]{article}
\usepackage{acl2015}
\usepackage{times}
\usepackage{url}
\usepackage{latexsym}

\usepackage{tikz-dependency}
\usepackage{hyperref}

\title{The Yahoo Query Treebank, V. 1.0\\
	\large May 2016}

\author{Yuval Pinter \\
  Yahoo Research \\
  {me@yuvalpinter.com} \\\And
  Roi Reichart \\
  Yahoo Research \& Technion IIT \\
  {roiri@ie.technion.ac.il} \\\And
  Idan Szpektor \\
  Yahoo Research \\
  {idan@yahoo-inc.com} \\}

\date{}

\begin{document}
\maketitle

\section{General}

This dataset release accompanies \newcite{Pinter2016} which describes the motivation and grammatical theory\footnote{This dataset is presented there in section 3.3.}. Please cite that paper when referencing the dataset.

The dataset may be accessed via the Yahoo Webscope homepage\footnote{\url{http://webscope.sandbox.yahoo.com}} under \textbf{Linguistic Data} as dataset \textbf{L-28}. The description in Section~\ref{sec:desc} is included within the dataset as a Readme.

The dataset is sure to have annotation errors which are not covered by the special cases specified in this document. Please approach the first author for any corrections and they will appear in the next release. See Section~\ref{sec:errors} for known errors.

\section{Dataset Description}
\label{sec:desc}

User queries annotated for syntactic dependency parsing, version 1.0.
These queries were issued on all search engines between 2012 and 2014 and led the searcher to click on a result link to a question page on the Yahoo Answers site\footnote{\url{http://answers.yahoo.com}}.

\subsection{Full description}

This dataset contains two files:\\

\begin{tabular}{|c|}
\hline
ydata-search-parsed-queries-dev-v1\_0.txt\\ 1,000 queries, 5,344 tokens\\
\hline
ydata-search-parsed-queries-test-v1\_0.txt\\ 4,000 queries, 26,015 tokens\\
\hline
\end{tabular}
\\\\
These files differ in their level of annotation, but share the schema. They contain tab-delimited lines, each representing a single token in a Web query. The tokens in each query are given sequentially, and queries are given in order of an arbitrarily-selected numeric ID (with no empty lines between queries). The field schema is detailed in Table~\ref{table:schema}.

\begin{table*}
\centering
\begin{tabular}{|l|l|}
\hline
Field \# & Field content\\
\hline
0 & query-id\\
1 & token-in-query (starting with 1)\\
2 & segmentation marker (`SEG' iff token starts a new segment, `-' otherwise)\\
3 & token form\\
4 & token part-of-speech $\dagger{}$\\
5 & index of syntactic head token, with 0 denoting root $\dagger{}$\\
6 & dependency relation of edge from head token to this token $\dagger{}\ddagger{}$\\
\hline
\end{tabular}
\caption{Fields marked by $\dagger{}$ are not populated for the dev set in V 1.0; fields with $\ddagger{}$ are not fully populated in the test set.}
\label{table:schema}
\end{table*}

\begin{table}
\centering
\begin{tabular}{|l|l|l|l|l|l|l|}
\hline
183 & 1 & SEG & charter & NN & 2 & nn\\
183 & 2 & - & school & NN & 0 & root\\
183 & 3 & SEG & graduate & VB & 0 & root\\
183 & 4 & - & early & RB & 3 & advmod\\
\hline
\end{tabular}
\caption{A sample query (\#{}183 in test set).}
\label{table:sample}
\end{table}

\begin{figure}
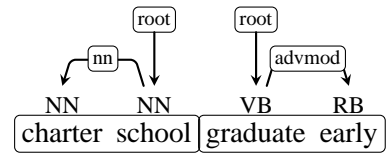

\centering
\begin{dependency}[edge style=thick]
   \begin{deptext}
      \small NN \& \small NN \& \small VB \& \small RB \\
      charter \& school \& graduate \& early \\
   \end{deptext}
   \depedge[edge unit distance=2ex]{2}{1}{nn}
   \deproot[edge unit distance=1.6ex]{2}{root}
   \deproot[edge unit distance=1.6ex]{3}{root}
   \depedge[edge unit distance=2ex]{3}{4}{advmod}
   \wordgroup{2}{1}{2}{pred}
   \wordgroup{2}{3}{4}{arg}
\end{dependency}
\caption{Query \#183 from the test set, tagged and parsed.}
\label{fig:example}
\end{figure}

An example query is brought in Table~\ref{table:sample}.
These lines represent a query, whose ID in the `test' set is 183, and whose raw form is \textit{charter school graduate early}. It is interpretable thus: the query is composed of two syntactic segments: \textit{[charter school] [graduate early]}. In the first segment, the syntactic root is the noun \textit{school}, and the noun \textit{charter} modifies it in a nominal compound modifier relation. The second segment is rooted by the verb \textit{graduate}, modified by the adverb \textit{early} in an adverbial modifier relation.
A dependency tree corresponding to this query is produced in Figure~\ref{fig:example}.

\subsection{Linguistic pre-processing notes}

All queries were tokenized using the ClearNLP tokenizer for English \cite{Clear} \footnote{Version 2.0.1, from \url{http://www.clearnlp.com}} and were not spell-corrected or filtered for adult content. For Excel-friendliness, initial quotes were replaced with backticks. When using off-the-shelf processing tools, we recommend to re-replace them in pre-processing.

\section{Annotation Guidelines}

The parse tree annotations followed the segmentation annotation. Both phases included supervision over automatic part-of-speech tagging.
In general, parse tree annotators were instructed to adhere to the Stanford Dependency guidelines \cite{sd2008}\footnote{Version 3.5.2, from \url{http://nlp.stanford.edu/software/dependencies_manual.pdf}} with the necessary caveats that come with non-standard text prone to errors.

Following are some selected issues which can be seen in the test dataset. Section~\ref{ssec:seg} describes issues relating to segmentation tagging, and the subsequent subsections address issues of dependency edge attachment.

\subsection{Segmentation Ambiguity}
\label{ssec:seg}

\paragraph{Noun Strings}
The most difficult segmentation decisions have been in cases of long strings of nouns, both common and proper. The guideline is based on judgment, asking whether \textbf{the phrase can stand as a constituent in a coherent sentence}. For example, in query 116 words 1-3 are \textit{big mac cost}, a phrase considered clunky as opposed to the much-preferred \textit{cost of (a) big mac}, resulting in the decision to segment it as \textit{[big mac] [cost]}. A more clear-cut case is the segmentation decision in q.284: \textit{[first day of missed period] [symptoms]}, where theoretically \textit{day} could be tagged as a modifier phrase head of \textit{symptoms}, but no imaginable well-formed sentence would use this formation as a constituent.

Sometimes reasonable semantics forced us to conclude towards a segmentation decision over an unlikely (but syntactically well-formed) single-segment constituent. For example, in q.271 \textit{[fanfiction] [guest reviews]} we assume (and support this decision by running our own search) that the searcher was looking for guest reviews on fanfiction website platforms rather than guest reviews written about pieces of fanfiction. Likewise, in q.3013 \textit{[questionnaire] [assisted suicide]} we treated \textit{assisted} as an adjective relating to \textit{suicide} rather than a main verb in a sentence describing an unlikely scenario.

\paragraph{Missing Auxiliary Verbs}
Another issue was certain constructions which may become a sentence by the addition of a copula or auxiliary verb. The guidelines stated that if the post-copular sequence describes an attribute of the semantic subject (pre-copula), it is considered the head of an adjunct phrase and shares its segment (e.g. q.2916 \textit{[paragraph \textbf{describing} a family]}, q.91.w.1-4: \textit{[battery cables not tight]}). If it is construed as a sentence missing an auxiliary, it is segmented away from the subject (e.g. q.475 \textit{[lab] [\textbf{throwing} up blood ?]}). In a way, this is an explicit application of the constituency test described earlier.

\subsection{Proper Names' Internal Structure}
The dataset contains many instances of product names followed by numbers denoting model (e.g. query 2347: \textit{how to replace crank sensor on \textbf{95 saab cs9000}}. The guidelines in these cases were to place the phrase head as the last alphabetic word (here, \textit{saab}) and depend the post-modifier (\textit{cs9000}) on it as an \textit{npadvmod}.

Long proper names with complex internal structures were left as their underlying structures. E.g. query 3252: \textit{the$_1$ call of the$_2$ wild} is parsed with \textit{call} as the phrase head with a \textit{det} preceding and a \textit{prep} following, further decomposed into \textit{pobj(of,wild)} and \textit{det(wild,the$_2$)}. However, more trivial internal structures were flattened into the common proper name representation: q.3260, \textit{jeeves \& wooster}: \textit{nn(wooster,jeeves)}, \textit{nn(wooster,\&)}. Nominal compounds were tagged as successive \textit{NNP}s, e.g. q.3413, \textit{what cameras do they use in \textbf{planet\_NNP earth\_NNP}}: \textit{nn(earth,planet)}.

\subsection{Truncated Sentences}
Many of the queries in the dataset are in fact sentences aborted mid-way. Some because the query is a sentence-completion question (e.g. query 970 \textit{a major result of the european age of exploration was}) and some for more opaque reasons (e.g. q.3828 \textit{why are russians so}). In both these types, where the root of the sentence would normally lie in the missing complement, the root was assigned to the token nearest to it in the assumed full graph (\textit{was} and \textit{so}, respectively), with the other dependents collapsed unto it. The same goes for any phrase which is truncated before its grammatical head (e.g. q.3840) or mandatory complement (e.g. q.1861).

\subsection{Foreign Languages}
The dataset contains several non-English queries that are treated as nonce (part-of-speech tag = `FW', dependency relation = `dep'). Their parse tree is, as a rule, a flat tree headed by the final word (proper name convention). E.g. q.3299, q.3319. Other cases where foreign words are tagged as `FW' is when they function meta-linguistically. E.g. q.3472, \textit{what does \textbf{baka\_FW} mean in japanese}.

\subsection{Grammatical Errors}
By the nature of Web Queries being written in real time by users possessing diverse proficiency and competence, the dataset contains many grammatical errors (as well as typos -- see Section~\ref{ssec:typos}). These were not corrected during pre-processing in order to maintain the authenticity of the data. The guidelines in such cases, unless re-segmentation was in order, called to retain as much of the intended structure as possible. For example, in query 3274, word 4 \textit{part} is meant to be \textit{parts} and as such is tagged as a plural noun. In some cases, different parts of the sentence were fused together due to incorrect grammar. The solution was to represent the fused token by the head of the intended phrase if it is fully contained within (e.g. q.3233, \textit{dogs} for \textit{dog 's}), or ignore a dependent if there is structural crossover (e.g. in q.3590, \textit{my sister in \textbf{laws} husband}, the possessive \textit{'s} was in effect excluded from the tree).

Another common case was auxiliary deletion common in ESL writing, e.g. query 3949 \textit{why you study in university}. Here we opted again for the intended meaning as a sentence missing the auxiliary \textit{do} rather than treating the full query as a constituent (adverbial clause, clausal complement, etc.) of a deleted governing predicate.

\subsection{Typographical Errors}
\label{ssec:typos}
In general, obvious typos were treated as the intended word whether they resulted in a legal English word or not. E.g. q.3 w.13, \textit{trianlgle\_NN}, or q.3786 w.5 \textit{if\_VBZ} (for the intended \textit{is}).\\
Sometimes, typos result in word merge, in which case they were either POS-tagged as \textit{XX} (e.g. q.252 w.3) or by the head if they can be construed as a coherent phrase (e.g. q.3115 w.1 \textit{searchwhere}, which is probably a mis-concatenation of a meta-linguistic \textit{search} with the first intended term \textit{where}, and so analyzed as if it were the latter alone).

Sometimes, extremely creative tokenization is employed by users. Behold the glory that is query 1814: \textit{green chemistry.in day today life}. We parsed \textit{day today} as if it were a noun phrase, and gave up representing the preposition \textit{in} altogether.

\subsection{BE-sentences}
Sentences with forms of the verb BE are often ambiguous between attributive sentences (where the BE-verb acts as copula to the head of the following phrase) and proper essential statements where BE is the main verb. We tended to go for the former in case of ambiguity, excepting for very clear cases of essence (e.g. query 3264 \textit{the free market is a myth}: \textit{root(ROOT,is)}) and, of course, where the following can only be a clausal or prepositional complement (e.g. q.799 \textit{trouble \textbf{is} you think you have time}, q.3593 \textit{what \textbf{is} on sheldons shirt in season 1 episode 4}).

\section{Known Annotation Errors}
\label{sec:errors}

\subsection{Segmentation Errors}
Test set:

\begin{tabular}{|l|l|l|}
\hline
Query ID & V 1.0 & Correct\\
\hline
194 & 1,3 & 1\\
304 & 1,3 & 1\\
325 & 1,3 & 1\\
362 & 1,3 & 1\\
425 & 1,4 & 1\\
847 & 1,3 & 1\\
911 & 1,4 & 1\\
3779 & 1,5 & 1\\
\hline
1147 & 1 & 1,3\\
1812 & 1 & 1,4\\
2784 & 1 & 1,2,3\\
2883 & 1 & 1,7\\
2912 & 1 & 1,5,7\\
3348 & 1 & 1,2\\
3366 & 1 & 1,2,3,4\\
\hline
\end{tabular}

\subsection{Attachment Errors}
Test set:

\begin{tabular}{|l|l|l|}
\hline
Query.Token & V 1.0 & Correct\\
\hline
3153.6 & 2 & 7\\
3153.7 & 6 & 2\\
\hline
\end{tabular}

\section*{Credits}

Segmentation tagged by \textbf{Bettina Bolla}, \textbf{Avihai Mejer}, \textbf{Yuval Pinter}, \textbf{Roi Reichart}, and \textbf{Idan Szpektor}. Parsing tagged by \textbf{Shir Givoni} and Yuval Pinter.

\bibliographystyle{acl}
\bibliography{bibliography.bib}

\end{document}